\documentclass[letterpaper]{article} 
\usepackage{aaai2026}  
\usepackage{times}  
\usepackage{helvet}  
\usepackage{courier}  
\usepackage[hyphens]{url}  
\usepackage{graphicx} 
\usepackage{natbib}  
\usepackage{caption} 
\usepackage{algorithm}
\usepackage{algorithmic}
\usepackage{algorithm}
\usepackage{algorithmic}

\usepackage{amssymb}
\usepackage{amsmath}

\usepackage{booktabs}
\usepackage{makecell}
\usepackage{multirow}
\usepackage{tabularx}
\usepackage{soul}
\usepackage{subcaption} 
\usepackage{siunitx}   

\usepackage{array}
\usepackage{calc}

\usepackage{newfloat}
\usepackage{listings}
\DeclareCaptionStyle{ruled}{labelfont=normalfont,labelsep=colon,strut=off} 
\floatstyle{ruled}
\newfloat{listing}{tb}{lst}{}
\floatname{listing}{Listing}
\nocopyright

\title{FFT-MoE: Efficient Federated Fine-Tuning for Foundation Models via Large-scale Sparse MoE under Heterogeneous Edge}
\author{
    Gang Hu, Yinglei Teng, Pengfei Wu, and Nan Wang,
}
\affiliations{
    \textsuperscript{\rm 1}Beijing Key Laboratory of Work Safety Intelligent Monitoring\\

    Beijing University of Posts and Telecommunications (BUPT), Xitucheng Road No.10, Beijing, China, 100876\\
}

\usepackage{bibentry}

\begin{document}

\maketitle

\begin{abstract}
As Foundation Models (FMs) drive progress toward Artificial General Intelligence (AGI), fine-tuning them under privacy and resource constraints has become increasingly critical—particularly when high-quality training data resides on distributed edge devices. Federated Learning (FL) offers a compelling solution through Federated Fine-Tuning (FFT), which enables collaborative model adaptation without sharing raw data. Recent approaches incorporate Parameter-Efficient Fine-Tuning (PEFT) techniques such as Low-Rank Adaptation (LoRA) to reduce computational overhead. However, LoRA-based FFT faces two major limitations in heterogeneous FL environments: structural incompatibility across clients with varying LoRA configurations, and limited adaptability to non-IID data distributions, which hinders convergence and generalization. To address these challenges, we propose FFT-MoE, a novel FFT framework that replaces LoRA with sparse Mixture-of-Experts (MoE) adapters. Each client trains a lightweight gating network to selectively activate a personalized subset of experts, enabling fine-grained adaptation to local resource budgets while preserving aggregation compatibility. To further combat the expert load imbalance caused by device and data heterogeneity, we introduce a heterogeneity-aware auxiliary loss that dynamically regularizes the routing distribution to ensure expert diversity and balanced utilization. Extensive experiments spanning both IID and non-IID conditions demonstrate that FFT-MoE consistently outperforms state-of-the-art FFT baselines in generalization performance and training efficiency.
\end{abstract}

\begin{links}
     \link{Code}{https://anonymous.4open.science/r/FMoE-9527}
\end{links}
\section{Introduction}
The rapid ascent of Foundation Models (FMs) is reshaping the landscape of artificial intelligence, enabling unprecedented generalization across diverse and complex tasks. These models have achieved remarkable success in domains such as computer vision \cite{FM4CV}, data mining \cite{FM4DM}, and natural language processing \cite{LLMSurvey}, largely due to their ability to learn rich and transferable representations from massive centralized datasets \cite{EfficientTrainingFMs}. However, this centralized training paradigm raises serious concerns about data privacy and regulatory compliance, especially in sensitive applications such as healthcare, finance, and personalized mobile services \cite{Fedprivacy}. To address these challenges,FL decentralizes model training by orchestrating gradient or model updates across clients, thus sidestepping the need for centralized data aggregation and preserving user privacy \cite{FedAvg}.
\begin{figure}[t]
\centering
\includegraphics[width=0.43\textwidth]{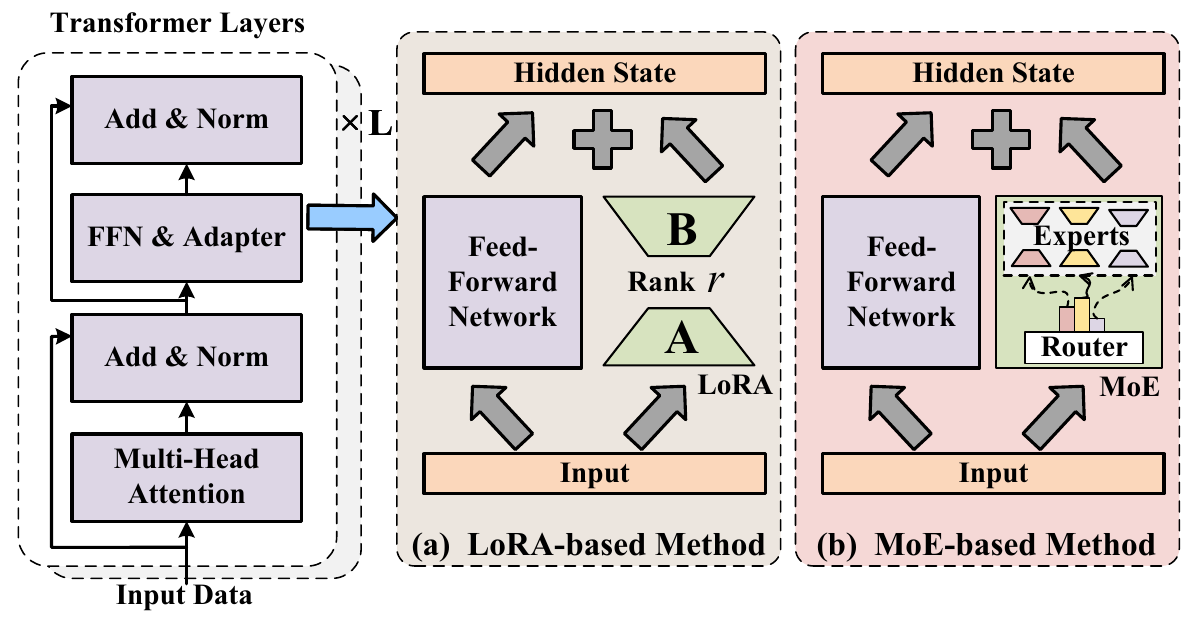} 
\caption{LoRA-Based Tuning vs. MoE-based Tuning.}
\label{MoEFineTuing}
\vspace{-5mm}
\end{figure}
While FL offers privacy advantages, directly training large-scale FMs in federated settings remains highly challenging due to the constrained communication bandwidth and limited computational resources on edge devices (e.g., training a GPT-3 model with 175B parameters will take seven months with 512 V100 GPUs. \cite{TrainGPT3}). To alleviate these limitations, Parameter-Efficient Fine-Tuning (PEFT) techniques, such as Low-Rank Adaptation (LoRA) \cite{LoRA}, enable clients to fine-tune and exchange only a small set of task-specific parameters while freezing the original model weights. This significantly reduces both the communication and memory overhead, making Federated Fine-Tuning (FFT)  viable on resource-constrained devices. 

However, applying FFT in real-world heterogeneous FL settings presents substantial challenges. \emph{On one hand}, device heterogeneity—the variation in computational capacity, memory, and network bandwidth across edge clients—complicates coordination and slows down training. Most FL systems rely on synchronous aggregation, where each round must wait for the slowest (straggler) device, significantly increasing overall training latency \cite{StragglerinFL}. \emph{On the other hand}, data heterogeneity, characterized by non-identically distributed data across clients,impedes convergence and often degrades global model accuracy \cite{Survey_Advance_and_Open_Problem_in_FL}. These challenges call for adaptive and resource-efficient fine-tuning strategies that can operate effectively and robustly under the diverse constraints of heterogeneous FL environments.

To mitigate these issues, recent efforts have explored integrating PEFT techniques into FFT frameworks. For example, FedPETuning  \cite{fedpetuning} evaluates several PEFT algorithms (e.g., adapter tuning, prompt tuning, and LoRA) within FL frameworks. Subsequent works have extended these approaches to better handle heterogeneity and improve efficiency. FedAdapter \cite{FedAdapter} dynamically configures the width and depth of inserted adapters to reduce training latency, while PromptFL \cite{FedPrompot} introduces soft prompts that enhance task adaptability across clients. Among PEFT techniques, LoRA has gained particular attention due to its effectiveness and zero additional inference overhead. FedIT \cite{FedIT} proposes a LoRA-based FFT approach and demonstrates its robustness on heterogeneous instruction-tuning datasets. FLoRA \cite{FLoRA} further extends this by supporting heterogeneous LoRA adapters via a stacking-based aggregation mechanism. FedPipePEFT \cite{FedPipePEFT} jointly optimizes LoRA configurations and batch sizes to reduce training costs without degrading inference performance. Despite these advancements, LoRA-based FFT still suffers from two  fundamental limitations in heterogeneous FL settings: (1) Structural incompatibility, where clients adopt different low-rank configurations, undermining synchronous aggregation; and (2) Limited adaptability to non-independent and identically distributed (non-IID) data, which hampers convergence and degrades global model generalization. These limitations highlight the need for a more flexible and heterogeneity-aware fine-tuning framework.

Unlike the previous works, we propose FFT-MoE a novel federated fine-tuning framework that leverages the Sparse Mixture-of-Experts (MoE) \cite{SwitchTransformers} as adaptive modules to address device and data heterogeneity in FL. In FFT-MoE, each client trains a lightweight local gating network to selectively activate a small subset of shared experts, enabling a dynamic trade-off between performance and resource efficiency. The MoE architecture’s expert-parallel design and sparse activation mechanism allow individual experts to specialize in different data distributions or tasks, which naturally supports fine-tuning under both non-IID data and device heterogeneity. However, sparse MoE architectures are prone to expert load imbalance  \cite{EfficientMoE}, where a few experts are overused while others remain underutilized, leading to degraded generalization and unstable training dynamics. 
To overcome this, FFT-MoE incorporates a heterogeneity-aware auxiliary loss that simultaneously promotes expert load balancing expert diversity and improving robustness to non-IID data distributions.
The key contributions of this paper are summarized as follows:
\begin{itemize}
    \item \emph{MoE-based Federated Fine-Tuning Framework:} We propose FFT-MoE, a novel FFT framework that integrates sparse MoE adapters to support efficient and flexible post-training of FMs under Heterogeneous FL setting.
    \item \emph{Scalable Expert Activation with Resource Adaptivity:} We design a personalized expert activation mechanism that allows each client to dynamically select the number of active experts based on local resource constraints, while maintaining structural compatibility for aggregation.
    \item \emph{Heterogeneity-aware Load Balancing:} We introduce a novel auxiliary loss that explicitly regularizes the routing distribution, mitigating expert overloading and underutilization. This promotes expert diversity and improves generalization under non-IID data distributions.
    \item \emph{Comprehensive Empirical Evaluation:} We conduct extensive experiments and ablation studies, demonstrating that FFT-MoE significantly outperforms FFT baselines in terms of convergence speed, generalization performance, and training efficiency.
\end{itemize}
\section{Background and Motivations}
To better present the methodology, we discuss closely related work here, focusing on federated fine-tuning on foundation models, 
and conduct the preliminary visualized load experiments of FFT-MoE.

\begin{figure*}[t]
\centering
\includegraphics[width=0.8\textwidth]{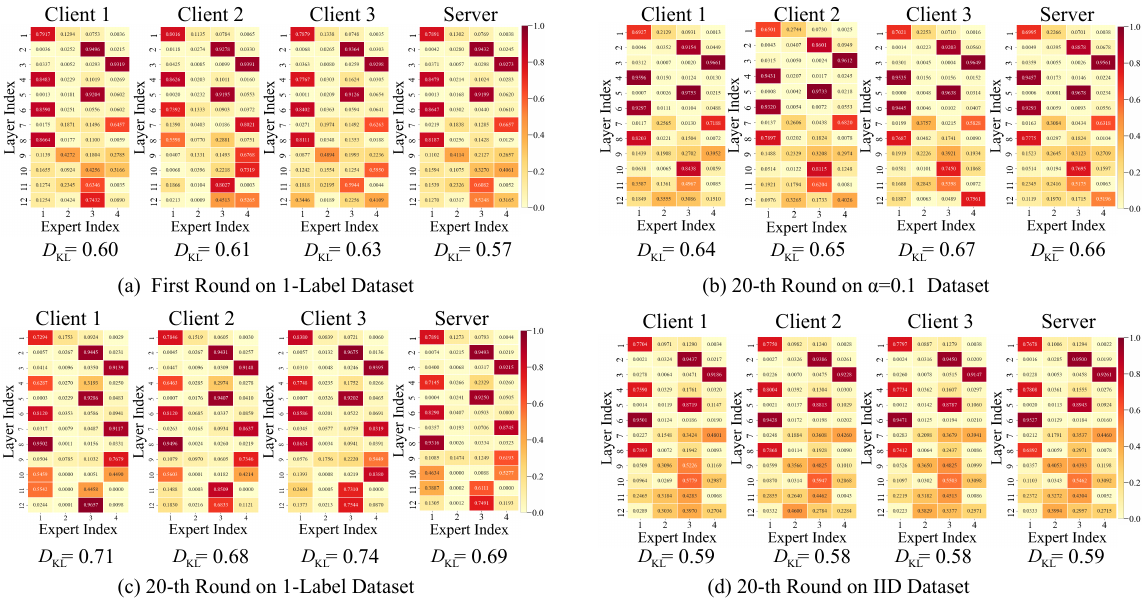} 
\caption{Expert load of local and global models on FL datasets. 
We apply the Sparse MoE mechanism to the BERT model and fine-tune it on the \emph{AgNews} dataset. To visualize expert utilization, we plot a heat map showing the activation distribution of experts across all transformer layers. In the heat map, darker colors indicate higher activation frequencies, implying that the corresponding experts are selected and updated more frequently during training. Conversely, lighter colors denote experts that are rarely updated.
${D_{{\text{KL}}}}$ is the average expert utilization of all layers based on KL divergence Eq. (\ref{Eq:KLDivergence}).}
\label{fig_ExpertLoading}
\vspace{-3mm}
\end{figure*}

\subsection{Injecting LoRA into Heterogeneous FFT}
Given the mismatch between the high computational demands of FMs and the limited resources of edge devices \cite{1}, integrating PEFT into FL serves as a practical alternative to full model training from scratch. As illustrated in Figure \ref{MoEFineTuing}, LoRA widely used PEFT method, fine-tunes pre-trained models by introducing low-rank projections \cite{2}. These projections are used to approximate the weight updates, allowing efficient fine-tuning of pre-trained models with minimal additional overhead. 


While LoRA-based FFT is effective on homogeneous device,  it fails to address critical challenges posed by data and device heterogeneity in FL.
For instance, heterogeneous FL requires flexible adapter sizes to accommodate devices with diverse resource constraints \cite{3}. High-capability clients can benefit from fine-tuning higher-rank adapters, while constrained devices require lower-rank ones. However, adapters with heterogeneous ranks lead to structural incompatibility, preventing direct model aggregation. Some existing methods \cite{FedPipePEFT,FLoRA} attempt to circumvent aggregation by merging low-rank projections into square matrices to align model shapes, but this transformation is irreversible and prevents decomposition back into original low-rank factors on local devices. Furthermore, vanilla LoRA fine-tuning does not inherently mitigate the impact of non-IID data distributions \cite{4}. The demand for adaptive computing are not unique to LoRA-Based FFT, but are shared across a broad class of current FFT methods \cite{5,6}. Hence, there is a pressing need for novel fine-tuning strategies that are not only parameter-efficient but also structurally and statistically adaptable to the heterogeneous nature of real-world FL systems.
\subsection{MoE is Antidote for Heterogeneity}
Federated fine-tuning systems are inherently challenged by various forms of heterogeneity—ranging from statistical non-IID data distributions to system-level disparities in computational capacity, memory, and communication bandwidth \cite{Survey_Advance_and_Open_Problem_in_FL}. These forms of heterogeneity often lead to unstable convergence, suboptimal generalization, and participation bias across clients. Sparse MoE architectures provide a principled mechanism to mitigate these issues by selectively activating a small subset of experts for each input \cite{7}, thus enabling partial specialization while maintaining global model sharing.
MoE offers two key advantages that make it  particularly well-suited for FFT: (1) its sparsity reduces per-client computation and communication overhead, facilitating deployment on resource-constrained devices \cite{8}; (2) its expert routing mechanism allows adaptive parameter utilization aligned with local data characteristics, thereby accommodating non-uniform data distributions across clients \cite{9}. Together, these strengths make MoE an ideal solution for heterogeneous FL environments, where personalized adaptation and communication efficiency are critical for practical deployment.

Motivated by these properties, we posit that MoE represents a scalable and architectural antidote to the core heterogeneity problems of FFT. To verify this viewpoint, we propose a novel MoE-based FFT framework which sparsely activates the trainable parameters by replacing the LoRA matrices with MoE module. Furthermore, we theoretically explore the intrinsic relationship between LoRA and MoE from the perspective of matrix decomposition. 
\begin{figure*}[t]
\centering
\includegraphics[width=0.95\textwidth]{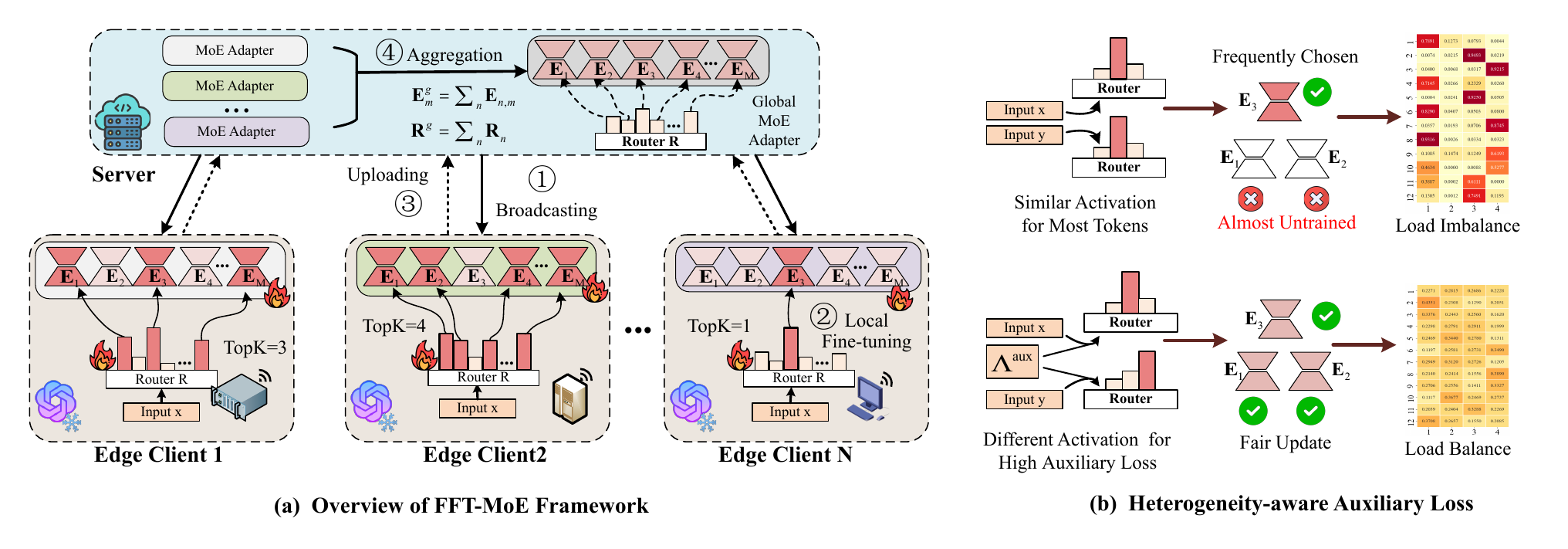} 
\caption{Left (a) is the pipeline of FFT-MoE, and right (b) is the illustration of auxiliary loss for expert load imbalance.}
\label{FFT_Framework}
\end{figure*}

While MoE-based architectures offer an elegant mechanism to alleviate system heterogeneity by assigning a variable number of experts per input token, they often introduce another challenge: \emph{expert load imbalance} \cite{SparseGatedMoE, 10}. In particular, the learned routing functions frequently over-utilize a small subset of experts across diverse clients and input distributions, leading to uneven parameter updates during training. This imbalance undermines the expected benefits of parameter sparsity by reducing the effective model capacity and compromises generalization and personalization in non-IID federated settings \cite{FedSMoE}. 

\subsection{Preliminary Experiments and Motivations}
In FFT scenarios, where local data distributions are highly skewed and expert activation is guided by local gradients, expert load imbalance becomes even more pronounced. Without proper coordination, certain experts may become heavily overloaded, while others remain under-trained, resulting in both reduced convergence stability and degraded downstream performance. Addressing this issue requires explicit load-balancing mechanisms that consider both local routing dynamics and global expert utilization.

To investigate the impact of expert load imbalance in heterogeneous FL, we conduct preliminary experiments under varying non-IID settings. The following insights inform the design of FFT-MoE:

\emph{1) Load imbalance correlates with model depth.}
As shown in Figure~\ref{fig_ExpertLoading}, shallow layers (e.g., layers 1–5) exhibit more pronounced expert imbalance than deeper layers (e.g., layers 9–11), as the latter capture task-specific semantics and activate more diverse experts. In contrast, shallow layers tend to develop dominant general-purpose experts due to limited representational diversity.

\emph{2) Expert preference is reinforced over training.}
At early rounds (e.g., round 1), expert load across clients is highly inconsistent in deeper layers due to uninformed routing. By round 20 (Figure~\ref{fig_ExpertLoading}(c)), local and global models exhibit more aligned expert usage, but this uniformity results from repeated aggregation and reinforces over-selection of certain experts, leaving others underutilized.

\emph{3) Data heterogeneity intensifies imbalance.}
As shown in Figures~\ref{fig_ExpertLoading}(b) and \ref{fig_ExpertLoading}(c), while moderate non-IID scenarios (e.g., Dirichlet partitioning with $\alpha$= 0.1) already induce some imbalance, extreme heterogeneity (e.g., single-label per client) results in severe load skew, particularly in shallow layers where representations are less specialized.  Under IID conditions, load distribution becomes more balanced. This confirms that increasing data heterogeneity amplifies both local and global load skew.

This imbalance intensifies across model depth, training rounds, and data distribution, leading to uneven expert utilization. As a consequence, some experts are overfitted while others remain under-trained, which negatively affects the updated fairness in under-sampling experts. This motivates our proposed solution, which incorporates adaptive expert routing and heterogeneity-aware regularization to improve efficiency and convergence in federated fine-tuning.

\section{The Proposed FFT-MoE Framework}
\subsection{Overall FFT-MoE Framework}
Consider a typical FFT framework consisting of a central server and a set of $N$ edge clients, each holding a local dataset $\mathcal{D}_n$. These distributed clients collaboratively optimize a global model ${\mathbf{W}_g} = \left[ {{{\mathbf{W}}^{{\text{FM}}}_g},{{\mathbf{W}}^{\text{E}}_g}} \right]$ by minimizing the following aggregated loss \cite{11}:
{\small
\begin{equation}
\label{Eq:FL_Loss}
{\mathcal{L}({{\mathbf{W}}_g}) = \sum\nolimits_{n = 1}^N {\frac{{|{\mathcal{D}_n}|}}{{\left| \mathcal{D} \right|}}{\mathcal{L}_n}({{\mathbf{W}}_n}|{\mathcal{D}_n})} ,}
\end{equation}
}
where $\mathcal{L}_n$ denotes the local loss function on local dataset $\mathcal{D}_n$ and $\mathcal{D} = \bigcup\nolimits_n {{\mathcal{D}_n}}$ represents the overall dataset of all clients.
The overall FFT-MoE framework is illustrated in Figure \ref{FFT_Framework}. In this framework, edge devices collaboratively fine-tune FMs over multiple rounds of FL using lightweight adapter modules, and each communication round consists of the following steps:
\begin{enumerate}
\item \emph{Initialization and adapters broadcasting:} All devices start with identical frozen pretrained weights $\mathbf{W}^{\text{FM}}$. The server broadcasts the global MoE parameters $\mathbf{W}_g^{\text{E}}$.
\item \emph{Local fine-tuning with Sparse MoE:} Each device independently fine-tunes its local MoE adapter $\mathbf{W}_n^{\text{E}}$ while keeping $\mathbf{W}^{\text{FM}}$ fixed. The local objective is to minimize the loss:
\begin{equation}\label{Eq:MoELoss}
    {\mathcal{L}_n(\mathbf{W}_n^{\text{E}} \mid \mathcal{D}_n, \mathbf{W}_n^{\text{FM}})},
\end{equation}
typically optimized using Adam or similar gradient-based optimizers.  
\item \emph{MoE adapters uploading:} After local fine-tuning, each device uploads its local MoE adapters $\mathbf{W}_n^{\text{E}}$ to the central server over the wireless communication link.
\item \emph{synchronous aggregation:} The server aggregates the local MoE adapters using a weighted average:
{\small
\begin{equation}\label{Eq:FL_Aggregation}
    {{\mathbf{W}}_g^{\text{E}} = \sum\nolimits_{n = 1}^N {\frac{{\left| {{\mathcal{D}_n}} \right|}}{{\left| \mathcal{D} \right|}}{\mathbf{W}}_n^{\text{E}}}}.
\end{equation}
}
\end{enumerate}
After aggregation, the updated global MoE module $\mathbf{W}_g^{\text{E}}$ is redistributed to all devices. This procedure is repeated over multiple communication rounds until convergence.
\subsection{Large-scale Sparse MoE with Adaptive Activation}
FFT-MoE can flexibly address personalized resource constraints by integrating the large-scale MoE with low rank and adaptively activation into adapter module. Specifically, in each Transformer block of the local frozen model ${\mathbf{W}}_n^{\text{FM}}$, we insert a trainable MoE adapter ${\mathbf{W}}_n^{\text{E}}$ into the feed-forward network (FFN). Each MoE adapter consists of:
\begin{itemize}
    \item A set of $M$ expert networks ${\mathbf{E}_1, \dots, \mathbf{E}_M}$, where each expert $\mathbf{E}_m: \mathbb{R}^d \rightarrow \mathbb{R}^d$ is a two-layer feed-forward network with intermediate dimension $r_m$.
    \item A learnable routing matrix $\mathbf{W}^R \in \mathbb{R}^{M \times d}$ used to compute routing logits.
\end{itemize}

Given an input token $x \in {\mathbb{R}^d}$, the output of the fine-tuning layer is:
\begin{equation}\label{Eq:SparseMoE}
    {x' = {{\mathbf{W}}^{{\text{FM}}}}x + \sum\nolimits_{m = 1}^M {{\mathbf{R}}{{\left( x \right)}_m}{{\mathbf{E}}_m}\left( x \right)}},
\end{equation}
where $\mathbf{R}(x)_m$ denotes the routing weight assigned to the $m$-th expert \cite{12}, and $\mathbf{E}_m(x)$ is the corresponding expert’s output. To enforce sparsity, only the top$K$ experts are selected:

\begin{equation}\label{Eq:SparseRouting}
    {{\mathbf{R}}\left( x \right) = {\text{Softmax}}\left( {\text{TopK}\left( {{{\mathbf{W}}^R}x} \right)} \right)},
\end{equation}
where $\text{TopK}(s)_i = s_i$ if $s_i \in \mathcal{S}_K(s)$ and $-\infty$ otherwise. Here, $\mathcal{S}_K(s)$ denotes the set of top-$K$ elements of $s$. 
To highlight the generality and expressive power of FFT-MoE, we demonstrate that LoRA can be interpreted as a special case of MoE with uniform expert activation.

\begin{figure}[t]
\centering
\includegraphics[width=0.35\textwidth]{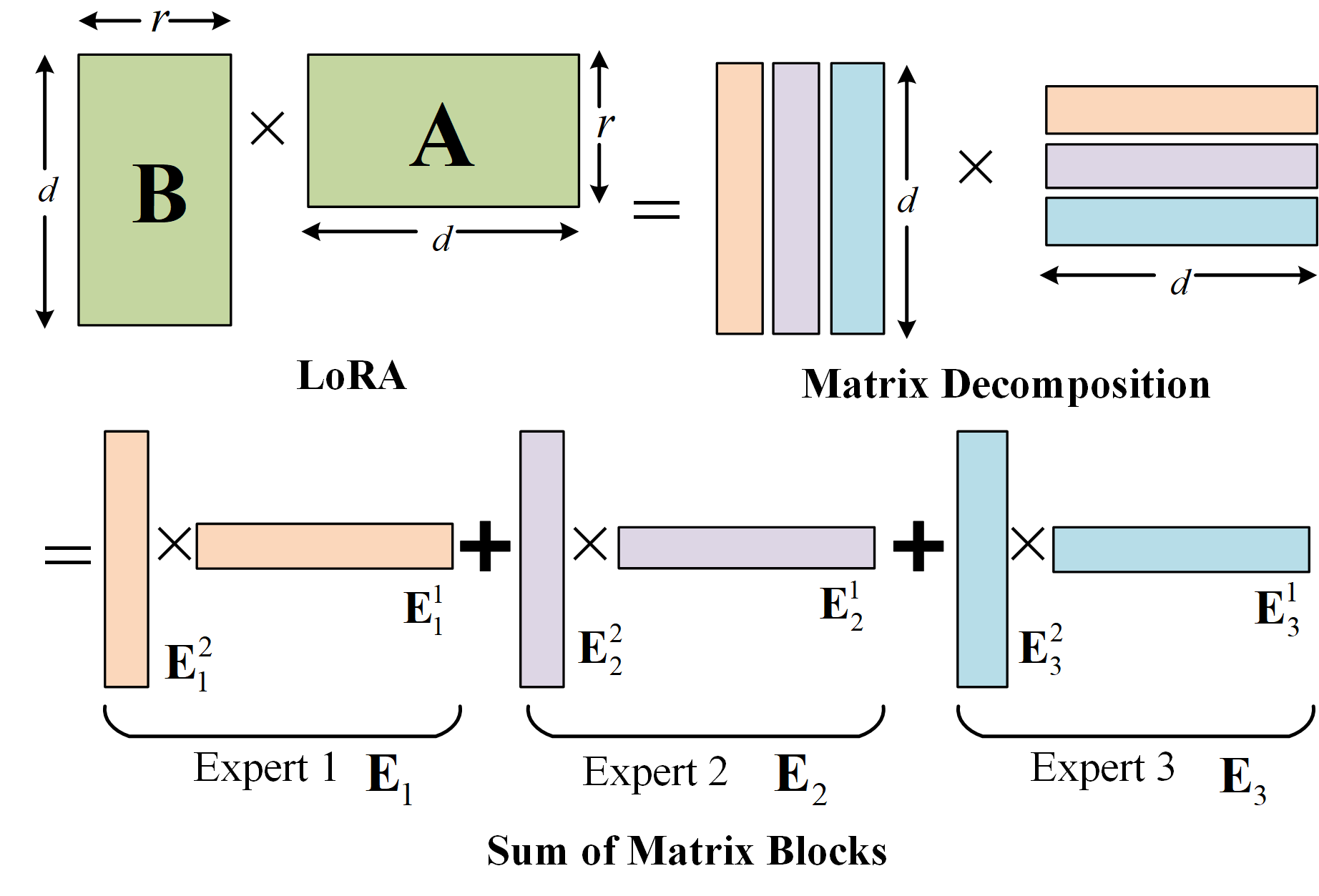} 
\caption{Illustration: LoRA as a special case of MoE.}
\label{MoEeqLaRA}
\vspace{-2mm}
\end{figure}

\subsubsection{LoRA as a special case of MoE with uniform activation}
In LoRA-based FFT, the updated weights of FMs are expressed as \cite{13}:
{\small
\begin{equation}
\label{Eq:LoRA_update}
{{\mathbf{W}} = {{\mathbf{W}}^{{\text{FM}}}} + \Delta {\mathbf{W}} = {{\mathbf{W}}^{{\text{FM}}}} + {\mathbf{BA}},}
\end{equation}
}where ${\mathbf{A}} \in {\mathbb{R}^{r \times d}}$, ${\mathbf B} \in {\mathbb{R}^{d' \times r}}$ and rank $r \ll \min \left( {d,d'} \right)$ are trainable low-rank matrices (with rank $r$).
$\mathbf{W}^{\text{FM}} \in \mathbb{R}^{d \times d'}$ represents the frozen pre-trained weights in a Transformer block, and $\mathbf{W}$ denotes the fine-tuned weights. 
Let $c_{i,j}$ denote the element at position $(i,j)$ in the update matrix $\mathbf{B}\mathbf{A}$:
{\small
\begin{equation}
{{c_{i,j}} = \sum\nolimits_{k = 1}^r {{{\mathbf{B}}_{i,k}}{{\mathbf{A}}_{k,j}}}}.
\end{equation}}where ${{\mathbf{A}}_{k,j}}$ is the element of ${\mathbf{A}}$ at position $(k,j)$. As illustrated as Figure. \ref{MoEeqLaRA}, the Matrix $\mathbf{A}$ can be split into multiple submatrices by rows and the Matrix $\mathbf{B}$ can be split into multiple submatrices by columns:
{\small
\begin{equation}\label{Eq:Rows}
    {{\mathbf{A}} = {\left[ {\mathop {{\mathbf{E}}_1^1}\nolimits^T ...\mathop {{\mathbf{E}}_m^1}\nolimits^T ...\mathop {{\mathbf{E}}_M^1}\nolimits^T } \right]^T}},
\end{equation}
\begin{equation}\label{Eq:Colums}
    {{\mathbf{B}} = \left[ {{\mathbf{E}}_1^2...{\mathbf{E}}_m^2...{\mathbf{E}}_M^2} \right]},
\end{equation}
}where $\mathbf{E}_m^1 \in {\mathbb{R}^{{r_m} \times d}}$, $\mathbf{E}_m^2 \in {\mathbb{R}^{d \times {r_m}}}$ and $\sum\nolimits_m {{r_m} = r} $. Define each expert as ${{\mathbf{E}}_m} = {\mathbf{E}}_m^2{\mathbf{E}}_m^1$. According to the multiplication rule of block matrices, the element $c_{i,j}$ can be equivalently written as the sum of the expert networks at the position $(i,j)$.
\begin{equation}\label{Eq:LoRAeqMoE}
{\small
    {\begin{gathered}
  {c_{i,j}} = \sum\nolimits_{k = 1}^{{r_1}} {{\mathbf{E}}_{1,i,k}^2{\mathbf{E}}_{1,k,j}^1}  + ,..., + \sum\nolimits_{k = 1}^{{r_{M}}} {{\mathbf{E}}_{M,i,k}^2{\mathbf{E}}_{M,k,j}^1}  \hfill \\
   = \sum\nolimits_{m = 1}^M {\sum\nolimits_{k = 1}^{{r_m}} {{\mathbf{E}}_{m,i,k}^2{\mathbf{E}}_{m,k,j}^1} }  = \sum\nolimits_{m = 1}^M {{{\mathbf{E}}_{m,i,j}}}.  \hfill \\ 
\end{gathered}}
}
\end{equation}
The above derivation shows that LoRA is a special case of MoE with uniformly activated experts, i.e., $K = M$ and $\mathbf{R}(x)_m = 1$. This equivalence suggests that FFT-MoE is capable of recovering LoRA behavior under certain routing constraints, ensuring no loss of performance in homogeneous settings.
\begin{table*}[htbp]
\centering
\setlength{\tabcolsep}{1.3mm}
\begin{tabular}{@{}l| c c c  c | c c c c@{}}
\toprule
\multirow{2}{*}{\makecell[c]{\textbf{Dataset} \\ \textbf{Model}}}
& \multirow{2}{*}{\makecell[c]{\textbf{Distribution}}} 
& \multirow{2}{*}{\makecell[c]{\textbf{FFT-} \\ \textbf{Adapter}}} 
& \multirow{2}{*}{\makecell[c]{\textbf{FFT-} \\ \textbf{Prompt}}} 
& \multirow{2}{*}{\makecell[c]{\textbf{FLoRA} \\ $R$=16}} 
& \multicolumn{2}{c}{\makecell[c]{\textbf{FFT-MoE}}} 
& \multicolumn{2}{c}{\makecell[c]{\textbf{FFT-MoE + Aux-Loss}}} \\
& & &  & & $R$=2, $M$=8 & $R$=4, $M$=4 & $R$=2, $M$=8 & $R$=4, $M$=4 \\
\toprule
\midrule
\multirow{5}{*}{\makecell[c]{\textbf{AgNews} \\ \textbf{BERT}}}
 & $1-$Label   & 43.90 {\small $\pm$3.85}& 45.13 {\small$\pm$2.5}& 61.92 {\small $\pm$2.53}&
 69.53 {\small $\pm$2.12}& 68.41 {\small $\pm$2.96}& \textbf{86.25} {\small $\pm$3.85}& 74.70{\small $\pm$3.02} \\
 & $\alpha =$ 0.1  & 50.56 {\small $\pm$1.19}& 53.86 {\small $\pm$1.34}& 88.38 {\small $\pm$1.21}
 & 89.53 {\small $\pm$0.98}& 88.70 {\small $\pm$1.54}& \textbf{89.87} {\small $\pm$2.14}& 89.63 {\small $\pm$0.87}\\
 & $\alpha =$ 1.0  & 72.11 {\small $\pm$2.2}& 87.34 {\small $\pm$1.05}& 93.46 {\small $\pm$3.87}
 & 92.96 {\small $\pm$0.83}& \textbf{93.92} {\small $\pm$2.76}& 93.13 {\small $\pm$1.34}& 93.34 {\small $\pm$1.78}\\
 & $\alpha =$ 10   & 74.66 {\small $\pm$1.78}& 88.08 {\small $\pm$0.64}& 93.78 {\small $\pm$0.98}
 & 93.29 {\small $\pm$1.15}& \textbf{94.38} {\small $\pm$0.40}& 93.25 {\small $\pm$0.26}& 93.35 {\small $\pm$0.53}\\
 & IID  & 74.61 {\small $\pm$0.65}& 88.31 {\small $\pm$0.53}& 93.65 {\small $\pm$0.65}
 & 93.24 {\small $\pm$1.12}& \textbf{94.34} {\small $\pm$0.19}& 93.18 {\small $\pm$0.65}& 93.37 {\small $\pm$0.64}\\

\midrule
\midrule

\multirow{5}{*}{\makecell[c]{\textbf{CIFAR-10} \\ \textbf{ViT}}}
 & $1-$Label        & 46.18 {\small $\pm$2.65}& 46.21 {\small $\pm$3.36}& 47.54 {\small $\pm$2.07}
 &80.46 {\small $\pm$2.12}&80.65 {\small $\pm$1.61}&\textbf{80.95} {\small $\pm$3.77}&80.80 {\small $\pm$2.35}\\
 & $\alpha =$ 0.1   & 48.89 {\small $\pm$4.68}& 75.09 {\small $\pm$1.74}& 79.05 {\small $\pm$2.02}
 &86.98 {\small $\pm$4.36}&87.39 {\small $\pm$2.91}&\textbf{88.19} {\small $\pm$4.18}&88.13 {\small $\pm$2.34}\\
 & $\alpha =$ 1.0   & 81.02 {\small $\pm$2.73}& 80.86 {\small $\pm$2.55}& 89.49 {\small $\pm$3.49}
 &\textbf{90.45} {\small $\pm$2.78}&90.43 {\small $\pm$1.56}&90.28 {\small $\pm$3.57}&90.43 {\small $\pm$1.07} \\
 & $\alpha =$ 10    & 82.57 {\small $\pm$2.08}& 85.83 {\small $\pm$2.63}& 89.89 {\small $\pm$3.97}
 &90.61 {\small $\pm$2.59}&\textbf{90.92} {\small $\pm$1.83}&90.61 {\small $\pm$2.58}&90.90 {\small $\pm$1.41}\\
 & IID              & 82.61 {\small $\pm$2.04}& 85.90 {\small $\pm$2.96}& 89.80 {\small $\pm$3.04}
 &90.32 {\small $\pm$2.77}&\textbf{90.86} {\small $\pm$1.59}&90.32 {\small $\pm$2.77}&90.84 {\small $\pm$0.89}\\

\bottomrule
\end{tabular}
\caption{Comparison of FFT-MoE with baselines on different datasets (in terms of accuracy \%), $R$ is the rank of experts.}
\label{Table:FFT-MoEPerfrmance}
\end{table*}

\begin{table}[htbp]
\centering
\setlength{\tabcolsep}{1mm}
\begin{tabular}{cccccccc}
\toprule
\multicolumn{2}{c}{\textbf{FLoRA}} & 
\multicolumn{2}{c}{\textbf{FFT-MoE}} & 
\multicolumn{1}{c}{\textbf{Aux-Loss}} &
\multicolumn{2}{c}{\textbf{FFT-MoE}} & 
\multicolumn{1}{c}{\textbf{Aux-Loss}} \\
$R$ & Accuracy &\multicolumn{2}{c}{$M$=2} & $M$=2 & \multicolumn{2}{c}{$M$=4}&$M$=4 \\
\midrule
4  & 86.49 & 2  & 88.21 &87.59 & 1 & 88.70 & 89.63 \\
8  & 87.67 & 4  & 88.29 &87.96 & 2 & \textbf{89.26} & 89.91 \\
16 & \textbf{88.38} & 8  & \textbf{89.63} &88.25 & 4 & 88.55 & \textbf{89.93} \\
32 & 88.21 & 16 & 88.61 & \textbf{88.67}& 8 & 87.51 & 89.54 \\
64 & 88.06 & 32 & 87.26 &88.33& 16 & 86.93 & 88.25 \\
\bottomrule
\end{tabular}
\caption{Tradeoff between rank and number of experts (\%)}
\label{tab:Comparison_rank_experts}
\end{table}
\subsubsection{Adaptive activation for heterogeneous computing} Unlike traditional Sparse MoE approaches that aim to train large models under a fixed resource budget, heterogeneous FL introduces a different challenge: clients, even with varying computational capacities, need to collaboratively train models of the same size. To address this, FFT-MoE allows each client to independently select its own sparsity level $K_n$, thereby enabling high-capacity devices to activate more experts while resource-constrained clients activate fewer. This adaptive activation strategy improves computational efficiency and scalability, and maintains the structural consistency of local models to perform federated aggregation.

\subsection{Heterogeneity-aware Auxiliary Loss}
While adaptive expert activation of FFT-MoE alleviates device heterogeneity, it introduces a new challenge: \emph{expert load imbalance}, further exacerbated by data heterogeneity. 
To address this issue, we propose a heterogeneity-aware auxiliary loss to regulate the expert selection and encourage balanced utilization across clients and layers under non-IID data. The auxiliary loss is integrated into the local training objective and dynamically applied based on the observed activation skew. The total training loss for client $n$ is defined as:
\begin{equation}
\label{Eq:TotalLoss}
{\mathcal{L}_n^{\text{total}} = \mathcal{L}_n^{\text{task}} + \lambda \cdot \mathcal{L}_n^{\text{aux}}},
\end{equation}
where $\mathcal{L}_n^{\text{task}}$ is the main task-specific loss (e.g., cross-entropy), $\mathcal{L}_n^{\text{aux}}$ is the proposed auxiliary term, and $\lambda$ is a weighting coefficient controlling the strength of regularization. 

Unlike traditional static balancing loss \cite{SwitchTransformers} applies uniform regularization regardless of the data distribution or model depth, our design adapts the constraint strength based on the real-time distribution of expert activation in the current data batch. When data heterogeneity is severe, stronger penalties are applied to encourage expert exploration; when data is more uniform, constraints are relaxed, especially in deeper layers where specialization may be beneficial.
The auxiliary loss into each layer $l$ is determined by the following expression:
{\small
\begin{equation}
\label{Eq:AuxLoss}
{\mathcal{L}_{n,l}^{\text{aux}} = \left\{ \begin{gathered}
  {D_{\text{KL}}}\left( {{\mathbf{P}_l}||{\mathbf{P}_g}} \right),\theta  \geqslant {\theta _{{\text{th}}}} \hfill \\
  0, \theta  < {\theta _{{\text{th}}}} \hfill \\ 
\end{gathered}  \right.},
\end{equation}
\begin{equation}
\label{Eq:KLDivergence}
{{D_{{\text{KL}}}}\left( {{{\mathbf{P}}_l}\left\| {{{\mathbf{P}}_g}} \right.} \right) = \sum\nolimits_{m = 1}^M {{{\mathbf{P}}_l}{{\left( x \right)}_m}\log \left( {\frac{{{{\mathbf{P}}_l}{{\left( x \right)}_m}}}{{{{\mathbf{P}}_g}{{\left( x \right)}_m}}}} \right)}},
\end{equation}
\begin{equation}
\label{Eq:ActivateThreshold}
{\theta  = \mathop {\max }\limits_m {\mathbf{P}}{\left( x \right)_m},}
\end{equation}}where ${{\mathbf{P}_l}{{\left( x \right)}_m}}$ denotes the logits output of routing network for expert $m$ and ${\mathbf{P}_g}$ is a uniform target distribution for global expert activation. ${D_{{\text{KL}}}}\left( {{{\mathbf{P}}_l}\left\| {{{\mathbf{P}}_g}} \right.} \right)$ is the KL-divergence \cite{14} between the local heterogeneity and global fairness. If $\theta$ exceeds the threshold $\theta _{{\text{th}}}$, indicating that the router is overly biased toward a single expert, the auxiliary term is activated. By minimizing this term, the router is encouraged to explore and utilize more experts which improves load balancing and parameter efficiency.

\section{Experiments}
In this section, we evaluate the performance of the proposed FFT-MoE framework under various heterogeneity settings on both language and vision tasks.

\subsection{Experimental Setup}

\subsubsection{Datasets and Models}
To validate the modality-agnostic design of FFT-MoE, we evaluate on two federated benchmarks:
\textbf{AgNews}\cite{Agnews}, a text classification dataset with 127.6K news articles across four topics; and
\textbf{CIFAR-10}\cite{CIFAR10}, an image classification dataset containing 60,000 32$\times$32 color images over 10 classes.
These datasets are chosen for their popularity in FL research and their diversity in both structure and modality.
For backbones, we use \textbf{BERT-base}\cite{bert} (12-layer, 110M parameters) for AgNews, and \textbf{Vision Transformer (ViT)}\cite{vit} (12-layer, 86M parameters) for CIFAR-10, both initialized with public pretrained weights.

\subsubsection{Heterogeneity Settings}
To simulate realistic FL scenarios, we consider both data and device heterogeneity:
\textbf{Data heterogeneity} is simulated by partitioning the datasets across clients using a Dirichlet distribution \cite{15} with varying concentration parameter $\alpha$. Smaller $\alpha$ values correspond to more skewed (non-IID) data distributions. Additionally, we include two special cases:
\emph{1-Label} (extreme non-IID): Each client receives data from only one class.
\emph{IID}: All clients are assigned uniformly random data samples.
\textbf{Device heterogeneity} is simulated by randomly assigning each client to one of two device types: high or low computational capacity, to reflect realistic variability in client-side resources.
\subsubsection{Baselines}

We compare FFT-MoE against three representative PEFT methods adapted for federated settings:
\textbf{FedAdapter}~\cite{FedAdapter}: Inserts lightweight adapter modules into intermediate model layers for efficient client-side adaptation.
\textbf{FedPrompt}~\cite{FedPrompot}: Prepends trainable soft prompt tokens to the input to guide the frozen backbone during downstream adaptation.
\textbf{FLoRA}~\cite{FLoRA}: Incorporates low-rank trainable matrices into the weight matrices of transformer layers to reduce communication and computation.
We run 5 times and report the mean and standard deviation of results.

\subsubsection{Parameter Settings and System Implementation}
We simulate a cross-device federated learning scenario with $N$=4 clients for Agnews and $N$=10 for CIFAR10, which all participate in each communication round. The global training is run for $T$=20 communication rounds. Each client performs one local epochs per round using a local batch size of 128. The optimizer used is Adam with a local learning rate of 3e-4 and weight decay of 0.01. All experiments are implemented in PyTorch and run on a distributed cluster of 8 NVIDIA A40 GPUs.


\subsection{Experiment Results}
\begin{figure}[t]
    \centering
    \begin{subfigure}[b]{0.21\textwidth}
        \includegraphics[width=\textwidth]{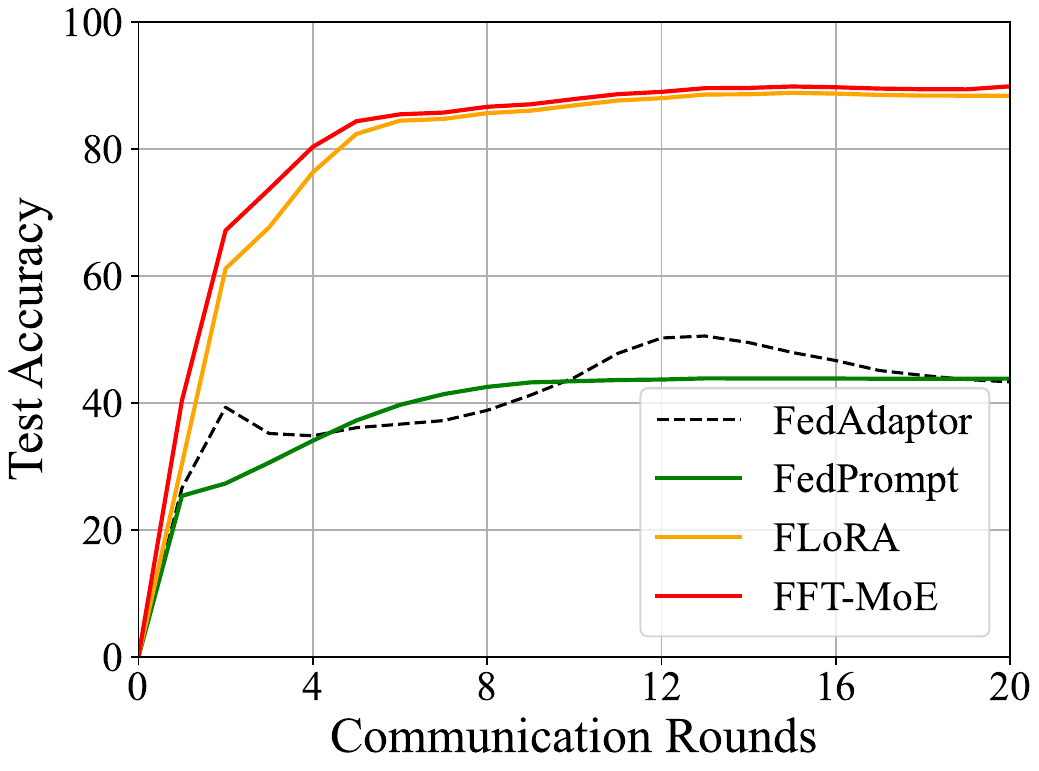}
        \caption{AgNews with $\alpha=$0.1}
        \label{fig:Acc-Bert-0.1}
    \end{subfigure}
    \begin{subfigure}[b]{0.21\textwidth}
        \includegraphics[width=\textwidth]{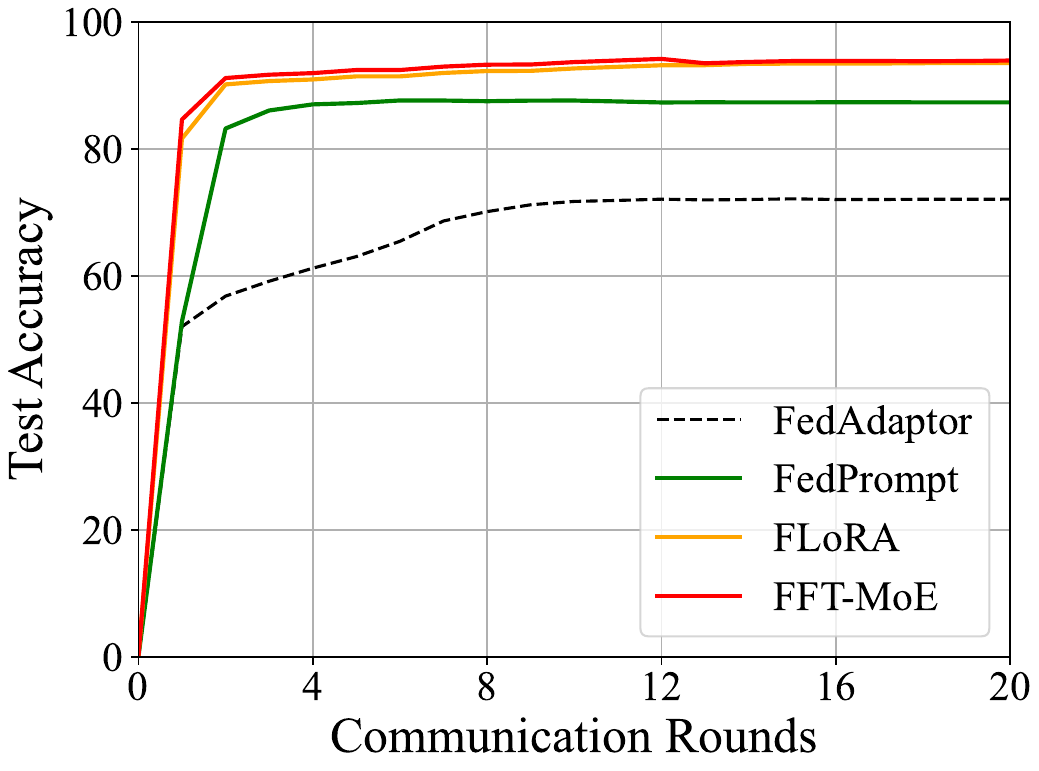}
        \caption{AgNews with $\alpha=$1.0}
        \label{fig:Acc-Bert-1.0}
    \end{subfigure}
    \\
        \begin{subfigure}[b]{0.21\textwidth}
        \includegraphics[width=\textwidth]{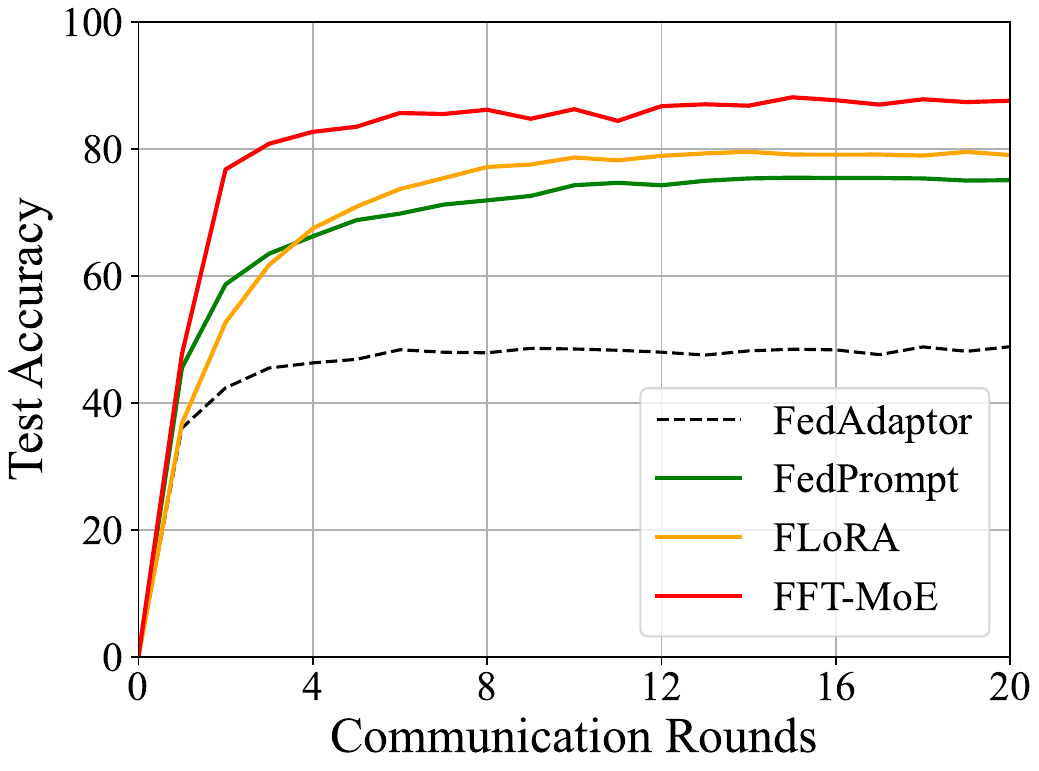}
        \caption{CIFAR10 with $\alpha=$0.1}
        \label{fig:Acc-ViT-0.1}
    \end{subfigure}
    \begin{subfigure}[b]{0.21\textwidth}
        \includegraphics[width=\textwidth]{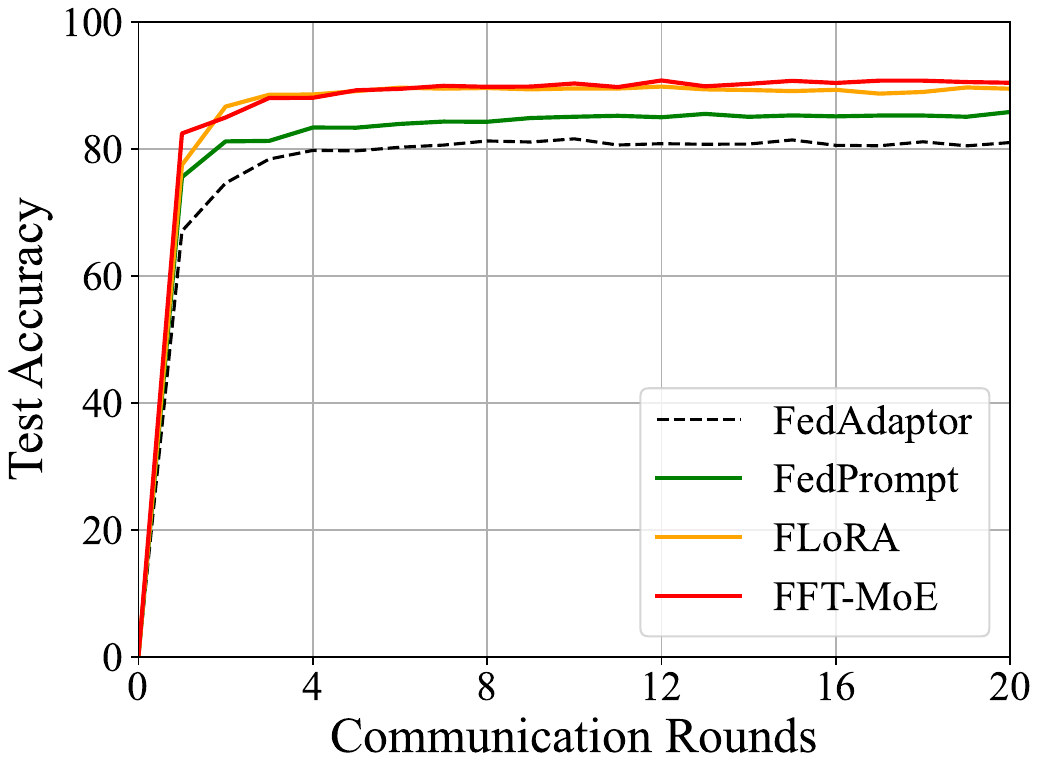}
        \caption{CIFAR10 with $\alpha=$1.0}
        \label{fig:Acc-ViT-1.0}
    \end{subfigure}
    \caption{The accuracy and convergence performance with different tuning methods}
    \label{fig:ACC-FFT-MoE}
\end{figure}

\begin{figure}[t]
    \centering
    \begin{subfigure}[b]{0.22\textwidth}
        \includegraphics[width=\textwidth]{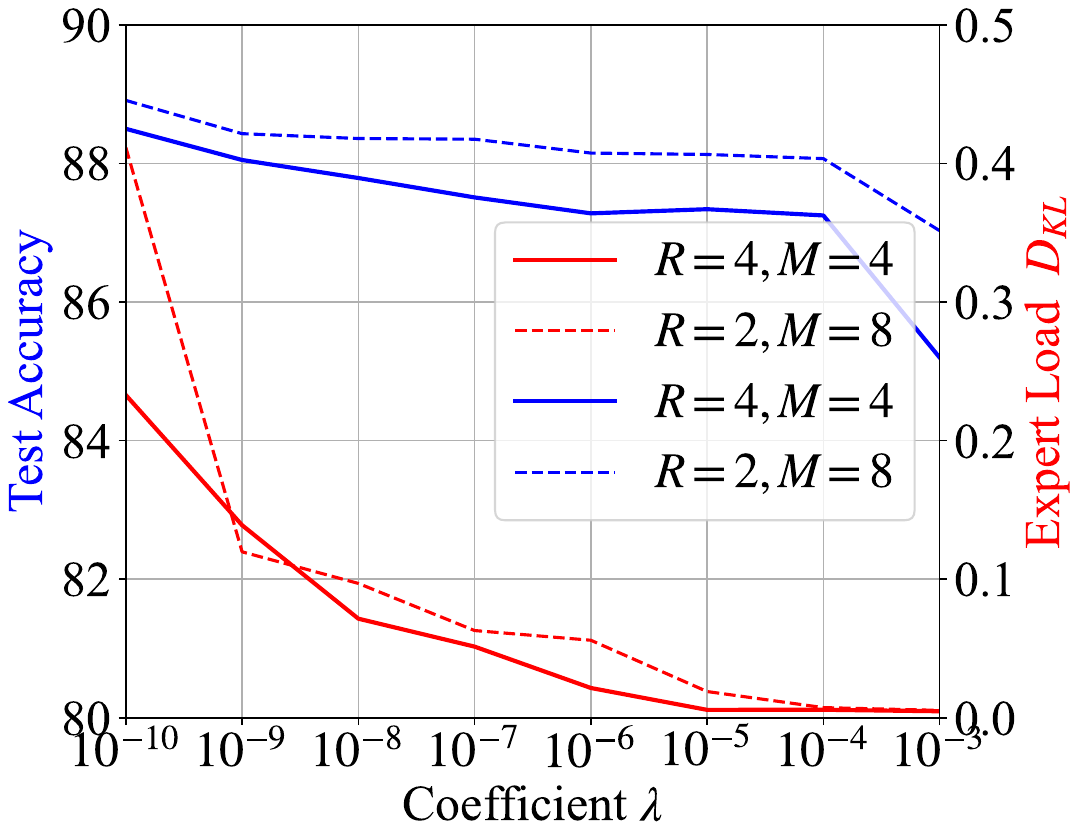}
        \caption{Weighting Coefficient $\lambda$}
        \label{fig:WeightedCofficient}
    \end{subfigure}
    \begin{subfigure}[b]{0.22\textwidth}
        \includegraphics[width=\textwidth]{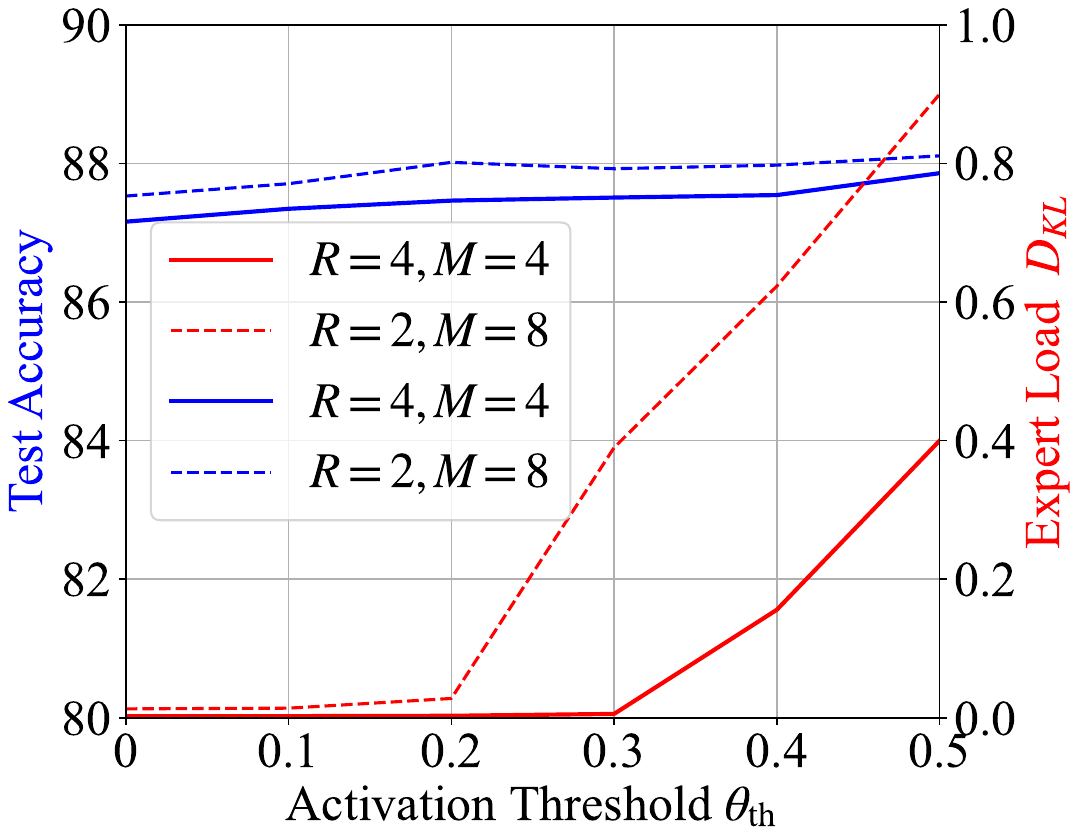}
        \caption{Activation Threshold $\theta_{\text{th}}$}
        \label{fig:Threshold}
    \end{subfigure}
    \caption{Effect of heterogeneity-aware auxiliary loss}
    \label{fig:AblationForLoss}
\end{figure}
\subsubsection{Performance of Proposed FFT-MoE}
Table~\ref{Table:FFT-MoEPerfrmance} presents the accuracy of FFT-MoE and baseline methods under varying degrees of data heterogeneity. All baselines suffer significant performance degradation as heterogeneity increases, particularly FFT-Adapter and FedPrompt, which lack the ability to dynamically route information or personalize effectively.
In contrast, FFT-MoE achieves consistently superior performance across all settings particularly in high non-IID degree.  On AgNews under the 1-label setting, the configuration ($R=2$, $M=8$) achieves 69.53\% accuracy, significantly outperforming FLoRA (61.92\%) and other methods. When incorporating our proposed auxiliary load-balancing loss, the accuracy further improves to 86.25\%, demonstrating a clear advantage in handling extreme data skew.
Notably, even under near-IID conditions, FFT-MoE remains competitive, achieving up to 94.34\% accuracy. This indicates that the expert routing mechanism does not overfit to heterogeneity but rather promotes robustness and generalization.

\subsubsection{Tradeoff Between Rank and Number of Experts}
Table~\ref{tab:Comparison_rank_experts} presents the test accuracy across various configurations of adaptation rank $R$ and number of experts $M$ for both FLoRA and FFT-MoE. A non-monotonic trend is observed: while increasing rank initially improves accuracy (e.g., FLoRA improves from 86.49\% at $R=4$ to 88.38\% at $R=16$), performance deteriorates beyond a certain point, indicating diminishing returns due to redundancy or instability. FFT-MoE exhibits a similar pattern.
Under the same parameter budgets, FFT-MoE with lower rank and more experts consistently outperforms higher-rank configurations. For example, FFT-MoE with $R=2$, $M=4$ achieves 89.26\% accuracy, surpassing FLoRA’s best result. Moreover, $R=2$, $M=8$ provides enhanced robustness under high data heterogeneity, likely due to its increased routing flexibility.
Incorporating the auxiliary load-balancing loss further improves performance, with FFT-MoE achieving a peak accuracy of 89.93\%. These results collectively suggest that: (1) increasing rank blindly is not always beneficial, and (2) given a fixed adaptation budget, FFT-MoE prefer lower rank and more expert configurations in non-IID data.

\subsubsection{Convergence of Proposed FFT-MoE}
To further compare the training dynamics across methods, we record the test accuracy versus communication rounds on different non-IID dataset. As shown in Figure~\ref{fig:ACC-FFT-MoE}(a), the FedPrompt and FedAdapter suffer from poor convergence and stagnate below 60\% accuracy. FLoRA improves performance but still converges slowly and plateaus early. In contrast, FFT-MoE quickly surpasses 80\% accuracy within 12 rounds, showing clear advantages in adaptation efficiency. Its accelerated convergence reflects the benefits of expert routing and reduced parameter conflict across non-IID clients. Under moderate heterogeneity ($\alpha$=1.0), Figure~\ref{fig:ACC-FFT-MoE}(b) and~\ref{fig:ACC-FFT-MoE}(d) shows FFT-MoE still maintains a margin over the baselines throughout training. These results confirm that \emph{the MoE architecture not only improves final performance but also accelerates optimization}, which is crucial for practical federated deployments with limited communication budgets.

\subsubsection{Ablation Study on Auxiliary Loss}
To quantify the impact of our heterogeneity-aware auxiliary loss, we conduct an ablation study on two hyperparameters: the loss weighting coefficient $\lambda$ and the expert activation threshold $\theta_{\text{th}}$.

\emph{Effect of weighting coefficient $\lambda$:} Figure~\ref{fig:AblationForLoss}(a) shows that small values (e.g., $10^{-10} \sim 10^{-8}$) have negligible impact on expert load balancing. However, extremely large $\lambda$ (e.g., $10^{-3}$) start to hurt performance due to over-regularization, which restricts expert specialization. Across settings, $\lambda = 10^{-5} \sim 10^{-4}$ offers the trade-off between diversity and specialization, consistently achieving the high accuracy and low load imbalance.

\emph{Effect of activation threshold $\theta_{\text{th}}$:} 
As shown in Figure~\ref{fig:AblationForLoss}(b), low threshold values (e.g., $\theta_{\text{th}} \leq 0.3$) result in balanced expert utilization, while higher thresholds deactivate the auxiliary term, increasing load imbalance. Importantly, accuracy remains robust across a wide threshold range. These results collectively suggest that: (1) increasing rank is not always beneficial, and (2) given a fixed adaptation budget, FFT-MoE prefer configurations with lower rank and more experts.

\section{Conclusion}
This paper introduces FFT-MoE, a novel federated fine-tuning framework that integrates sparse MoE with parameter-efficient techniques to address the dual challenges of device and data heterogeneity in federated learning. By leveraging conditional expert activation, FFT-MoE enables flexible adaptation to varying client capabilities, which serves as an effective method to address device heterogeneity. To mitigate the expert load imbalance exacerbated by data heterogeneity, we further propose an auxiliary balancing loss that encourages expert diversity without sacrificing specialization.
The experiment results show that FFT-MoE consistently outperforms other baselines which achieves better accuracy and convergence stability. Furthermore, we demonstrate that distributing adaptation capacity across multiple lightweight experts yields better robustness and performance than increasing rank alone.

\bibliography{aaai2026}


\end{document}